# Human Action Recognition (HAR) Using Skeleton-based Spatial Temporal Relative Transformer Network: ST-RTR

Faisal Mehmood[1], Enqing Chen[1,2*], Touqeer Abbas[3], and Samah M. Alzanin[4]

[1]School of Electrical and Information Engineering, Zhengzhou University, Zhengzhou, 450001, Henan, China.

[2]Henan Xintong Intelligent IOT Co., Ltd., Zhengzhou 450007, Henan, China.

[3]Department of Computer Science and Technology, Beijing University of Chemical Technology, Beijing, 100029, China.

[4]Department of Computer Science, College of Computer Engineering and Sciences, Prince Sattam bin Abdulaziz University, Al-Kharj, Saudi Arabia.

Corresponding Authors: ieeqchen@zzu.edu.cn

## Abstract

Human Action Recognition (HAR) is an interesting research area in human-computer interaction used to monitor the activities of elderly and disabled individuals affected by physical and mental health. In the recent era, skeleton-based HAR has received much attention because skeleton data has shown that it can handle changes in striking, body size, camera views, and complex backgrounds. One key characteristic of ST-GCN is automatically learning spatial and temporal patterns from skeleton sequences. It has some limitations, as this method only works for short-range correlation due to its limited receptive field. Consequently, understanding human action requires long-range interconnection. To address this issue, we developed a spatial-temporal relative transformer ST-RTR model. The ST-RTR includes joint and relay nodes, which allow efficient communication and data transmission within the network. These nodes help to break the inherent spatial and temporal skeleton topologies, which enables the model to understand long-range human action better. Furthermore, we combine ST-RTR with a fusion model for further performance improvements. To assess the performance of the ST-RTR method, we conducted experiments on three skeleton-based HAR benchmarks: NTU RGB+D 60, NTU RGB+D 120, and UAV-Human. It boosted CS and CV by 2.11 % and 1.45% on NTU RGB+D 60, 1.25% and 1.05% on NTU RGB+D 120. On UAV-Human datasets, accuracy improved by 2.54%. The experimental outcomes explain that the proposed ST-RTR model significantly improves action recognition associated with the standard ST-GCN method.

# 1. Introduction

In recent years, people have become more interested in HAR due to the progress made in deep learning and how it can be used in human-computer interactions, care for the elderly and disabled people in healthcare, and video surveillance [1, 2]. Despite this, many questions about skeleton-based action recognition need to be answered.

Nowadays, Graph Neural Networks (GNNs), and particularly Graph Convolutional Networks (GCNs), are the most broadly proposed technique for skeleton-based HAR because they effectively represent non-Euclidean data and can successfully capture both spatial (intra-frame) and temporal (inter-frame) information. S. Yan *et al.* [3] were the first to introduce GCN models for skeleton-based HAR. These models are often named Spatial Temporal-Graph Convolutional Networks (ST-GCNs). These models handle spatial data by manipulating the spatial links between skeleton joints taking temporal data by considering the temporal relations between each skeleton joint. Although ST-GCN models have been shown to perform exceptionally well on skeletal data, they have significant structural limitations addressed by other researchers, such as [4-6].

This research introduced a new mechanism on a light transformer, called a relative transformer, that eliminates the issues identified in the prior study. Using a relative transformer mechanism, the signals are sent in a spatial skeleton-based architecture by establishing a link between two independent joints. The spatial-temporal dimension also captures intra-frame and inter-frame interactions between two distant frames. Consequently, we call this model the "spatial-temporal relative transformer" (ST-RTR). So, we offer a fusion model that efficiently expresses an ST-RTR skeleton sequence's output.

- First, the human body graph's topology is fixed for all layers and activities. It may be difficult to create comprehensive visualizations of how the skeleton moves over time if graph linkages are straight and information travels beside the specified way.
- Second, A standard 2D convolution is used for both spatial and temporal convolution. As a result, they can only work nearby, which is limited by the convolution kernel's size.
- Third, when clapping, it's easy to overlook the significance of kinematic similarities between opposing body parts, such as the left and right hands.

In this study, we used a modified relative transformer module to overcome all these issues, shown in Fig. 1. Relative Transformer has shown itself to be highly effective on many different Computer Vision (CV) tasks, despite its initial development for NLP applications. Now, various transformer-based models have been introduced to enhance the effectiveness of the original idea, including the Set Transformer [7], the Routing Transformer [8], and the Star Transformer [9]. Many of these models aim to reduce computational complexity and memory requirements. Transformers were designed for NLP and CV. The first object detection system that incorporated a transformer and a CNN was the Detection Transformer (DETR) [10]. Vision Transformer (ViT) [11] uses transformer architecture instead of a standard CNN for image classification tasks and brings state-of-the-art consequences. Our study model consists exclusively of transformer-related components.

The Star Transformer and Skeleton Transformer stimulated the proposed Relative Transformer; these transformers were developed for skeletal data-based HAR.

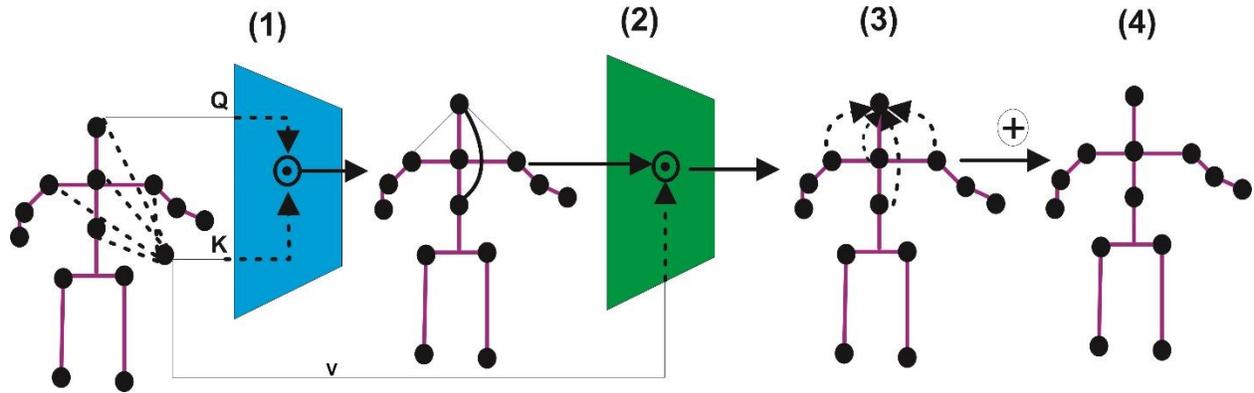

**Fig. 1.** Relative Transformer for skeleton joints. (1) Query vector (Q), key vector (K), and value vector (V) are calculated for each body joint. (2) The dot product (⊙) of each joint's query and key vectors represents its connection strength to all other joints. (3) Finally, each node's connection with the current node scales it. (4) New attributes are summed from the weighted nodes.

## 1.1 Contributions and advantages of the proposed model

This research presents a new mechanism, the Spatial-Temporal Relative Transformer (ST-RTR), to overcome the limitations of existing Graph Convolutional Networks (GCNs), specifically ST-GCNs, for skeleton-based HAR. The ST-RTR utilizes a modified relative transformer module to address issues such as fixed human body graph topology, limited spatial and temporal convolution, and overlooking kinematic similarities between opposing body parts.

The Spatial-Temporal Relative Transformer (ST-RTR) provides several advantages over existing methods for skeleton-based HAR. First, the ST-RTR overcomes the limitations of fixed human body graph topology in Graph Convolutional Networks (GCNs), such as ST-GCNs. The ST-RTR uses a modified relative transformer module that establishes a link between two independent joints, allowing for more comprehensive visualizations of how the skeleton moves over time. Second, the ST-RTR overcomes the limitations of limited spatial and temporal convolution in existing methods. It utilizes a transformer-based architecture that can efficiently capture spatial (intra-frame) and temporal (inter-frame) interactions between distant frames. This approach provides a more effective way to represent and process skeletal data. Third, the ST-RTR considers the kinematic similarities between opposing body parts, such as the left and right hands. This approach allows the model to capture more critical information about the motion of the human body.

Our research findings are summarized below:

➢ Our novel two-stream transformer model recognizes skeletal action using spatial and temporal relative transformers.

➢ For spatial and temporal modeling, a lightweight relative transformer model is developed. We present a spatial relative transformer module (S-RTR) that operates in the spatial dimension to construct long-range dependencies while keeping the basis skeleton topology.

The T-RTR model evaluates the connection between non-consecutive frames for more time without modifying the skeleton sequence.

➢ ST-GCN model takes skeleton joints from the ST-RTR stream as input, and the spatial-temporal relative transformer computes features as outputs in the fusion model.

➢ With joint and bone information, our model consistently outperformed ST-GCN [11], A-GCN [12] and MSST-RT [13], achieving state-of-the-art consequences.

The rest of the study is organized as follows: We describe related work in section 2. The background of this research is demonstrated in section 3. Implementation of the proposed is designed in section 4. Section 5 explains the model Evaluation. Finally, section 6 presents the conclusion.

## 2. Related Work

### 2.1 Skeleton-based HAR

Most of the earliest models used human-created characteristics to identify actions from their bone structures [14, 15]. Deep learning revolutionized activity recognition by providing practical approaches to increasing resilience and achieving remarkable performance [16, 17]. This class of methods uses several skeleton-related data: (1) A RNN [16, 17] technique leverages a sequential relationship between two points, considering the non-stationary insertion of musculoskeletal information. (2) Methods based on conventional neural networks CNNs [18, 19] use Geospatial and RNN-based data. A pseudo-image shows temporal dynamics and skeleton joints from a three-dimensional bone sequence in rows and columns. (3) The natural topological graph structure of the human skeleton is used in the graph neural network (GNN) based approaches [5, 20] that made utilization of both spatial and temporal data. Among the three techniques, the ST-GCN is the most expressive and the first to capture the balance between spatial and temporal relationships [21, 22]. The ST-GCN baseline model was employed in this study, and the workings of this model are described in depth in Section 3.2. We propose using the switcher self-attention operator instead of standard graph convolutions for spatial and temporal. S. Cho *et al.* [23] presented SAN, in which the action sequence is segmented into temporal clips, and self-attention is employed to represent long-term semantic data. Therefore, this method does not consider the correlations between low-level joints inside and across frames since it focuses on course-grained embedding instead of skeleton joints. Instead, we effectively describe the skeleton sequence's spatial and temporal relationships by fully realizing directly the graph's intra and inter-frame nodes.

### 2.2. Graph Neural Networks (GCNs)

Deep learning models may be extended to non-Euclidean domains using geometrical deep convolutional neural networks [24]. To begin with, M. Gori *et al.* [25] introduced the concept of the GNNs, which was further developed by [26]. In GNNs, nodes represent ideas or things, and edges reflect their connections. This simple notion is at the heart of GNNs. The development of CNNs made it possible to generalize convolution from grid data to graph data. Each time a GNN repeatedly processes the graph, it represents each node because it modifies vertices and their neighbor's characteristics. J. Bruna *et al.* [27] adapted convolution to signals by employing a spectral architecture in their first statement of CNNs on graphs. Graph-based CNNs were initially proposed by [27, 28], who utilized a spectral architecture to extend convolution to signals. T. N. Kipf and M. Welling [29] have simplified and expanded on this (2017). Graph convolution is a supplementary method for information aggregation in a spatial context [30]. The spectral architecture developed by [29] is used in Section 3.2 of this study.

### 2.3. Transformers Applied to Computer Vision

A. Vaswani *et al.* [30] introduced the Transformer as an alternative to RNNs and the leading approach for NLP. The production process for cinematic intervals involves generating a sequence of words or phrases closely related to a given input. This can be difficult for LSTMs and RNNs to handle because these models are designed to process data sequences, but they may struggle when the lines are very long or complex. It has been developed to meet these challenges. According to A. Vaswani *et al.* [30], while the Transformer utilizes the conventional encoder-decoder construction, it depends entirely on multi-head self-attention. The Transformer self-attention algorithm has been used in various well-known CV applications in recent years. A discrete non-local operator based on self-attention was presented by [31]; this operator captures spatial-temporal dependencies to improve video classification. A. Dosovitskiy *et al.* [32] suggested a Vision Transformer (ViT), which demonstrates how Transformers may successfully switch regular convolutions on pictures, as a follow-up to the effort to employ self-attention as an alternative to convolutional operators [33]. I. Bello *et al.* [33] created an original image transformer architecture to solve the picture captioning challenge. The initial effort to employ a Transformer model address to find the issues was undertaken by [10], and the model they used was called the Detection Transformer (DeTR). Point Transformer is a model suggested by [34]. This model converts self-attention to express connections between point clouds, capitalizing on the permutation-invariant nature of point clouds. Numerous different Converters software in object segmentation [35, 36] and generative modeling [37, 38] have recently been developed, demonstrating the possibility of transformer models being used for various tasks.

### 2.4. Research Gaps

The research presented in this paper aims to improve the performance of skeleton-based action recognition by addressing the limitations of existing methods, such as the narrow receptive field of graph convolutional networks and the inability of traditional transformers to capture relative positional information. We propose a novel approach that uses relative transformers to capture the spatial and temporal relationships in the skeleton data, preserving the graph's topology while requiring less complexity due to the graph's unordered sequence. We present experimental results demonstrating the effectiveness of the ST-RTR model on four benchmark datasets: NTU RGB+D 60, NTU RGB+D 120, and UAV-Human. The results show that the ST-RTR model outperforms several state-of-the-art methods, including the ST-GCN, and achieves higher accuracy while requiring fewer parameters.

# 3. Background

This part summarizes the main building elements of the proposed model in this paper. These building blocks are the geographic ST-GCN by [3] and the classic Relative Transformer MSST-RT [13], respectively.

### 3.1 Skeleton Sequences Representation

The number of joints constituting each skeleton is denoted by the variable *V*. The variable *T* represents the number of skeletons that make up the series. In follows, we will refer to the individual skeletons that make up the sequence as frames. A spatial-temporal graph is constructed to describe in series, for example, $G = (N, E)$, where $N = \{v_{tx} \mid t = 1, \ldots T, = 1, \ldots V\}$. The nodes $v_{tx}$ in the graph is the skeleton's joints over time, are represented by *E,* a set containing all the interconnections between them. *E* is divided into two parts, the first of which is $E_s = \{(v_{tx}, v_{ty}) \mid x, y = 1, \ldots V, t = 1, \ldots T\}$. During each time interval *t*, the intra-frame connections between each pair of joints *(x, y)* linked by one or more bones in the human bone structure. Using some criteria, the subset *ES* network of intra-frame connections is usually further split into *K* discrete pieces [3] (for example, the time travel from the origin of gravity), utilizing an adjacency matrix, and then encoded $A \in \{1, 0\}^{v \times v}$. The second subset $E_T = \{(v_{tx}, v(t+1)x \mid x = 1, V, t = 1, T\}$ is inter-frame linkages between joints that span many periods. A graph spanning the geographical and temporal dimensions is created [13].

### 3.2. Spatial-Temporal Graph Convolutional Networks (ST-GCN)

S. Yan *et al.* [3] presented the concept of ST-GCN. The ST-GCNs are made up of stacked blocks of space and time. Each block has a temporal convolution (TCN) which occurs after a spatial convolution (GCN). According to T. N. Kipf and M. Welling [29], the spatial sub-module employs the following GCN formulation:

$$f_{out} = \sum_{k}^{k_s} (f_{in} A_k) W_k. \tag{1}$$

$$A_k = D^{-\frac{1}{2}} (A_k + I) D_k^{-\frac{1}{2}}, D_{xx} = \sum_{k}^{k_s} (A_k^{xy} + I_{xy}) \tag{2}$$

where $K_s$, the spatial dimension, is the kernel size. In an undirected graph, the adjacency matrix $A_k$ represents the intra-frame connections. In this case, $W_k$ is a weight matrix that can be trained, and *I* is the independent variable. The temporal convolution network, or *TCN*, is a function that processes inputs of size ($C_{in}$, *V, T*) by considering $K_t$ frames at a time and producing an output of size *(V, T)*. Machine learning and computer vision often use it to analyze data sequences over time. The *TCN* operates on the *(V, T)* input dimensions, which are spatial and temporal. The receptive field is being evaluated. Because the adjacency matrix is unchangeable, the graph's structure is already determined, as shown by Eq. (1). To make it flexible, L. Shi *et al.* [4] developed an A-

GCN. In this network, the formulation of the GCN that is found in Eq. (1) is replaced with the following:

$$f_{out} = \sum_{k}^{k_v} f_{in}(A_k + B_k + C_k)W_k \qquad (3)$$

where $A_k$ likewise in Eq. (1), Training provides $B_k$ and Using a similarity function, and $C_k$ evaluates if two vertices are associated.

### 3.3. Relative Transformer

The standard Transformer model proposed by [30] contains self-attention, a non-local operator meant to improve the insertion of each word depending on its nearby context. To make it more word-dependent, the Transformer compares two words and then mixes them based on how relevant each word is to the other. The relative transformer allows one to obtain a better meaning from each word by acquiring cues from the context around them. The standard Transformer uses a set of positional encodings $U \in R^{L_{max} \times d}$, where the $i^{th}$ row $U_i$ represents the $i^{th}$ *absolute* location inside a segment, and $L_{max}$ represents the maximum length. The word embeddings and positional encodings are added element by element as the input to the Transformer. By applying positional encoding to our recurrence mechanism, we can calculate the hidden state sequence using the following process:

$$h_{\tau+1} = f(h_\tau, E_{s_{\tau+1}} + U_{1:L}) \qquad (4)$$

$$h_\tau = f(h_{\tau-1}, E_{s_\tau} + U_{1:L}) \qquad (5)$$

where $E_{s_\tau} \in R^{L \times d}$ is the word depending on the sequence of $s_\tau$, a transformation function is $f$. $E_{s_\tau}$ and $E_{s_{\tau}+1}$ are connected similar positional encoding $U_{1:L}$. The model has difficulty distinguishing between positions in a sequence because it lacks information about their relative positions. To fix this, the model can use "relative positional encoding," which gives the model a temporal bias or "clue" about how to gather data and where to focus. This can be done by inserting the relative distance between positions into the attention score, allowing the model to distinguish between positions based on their close distances. This method has been previously explored in other contexts. Still, the authors present a new form of relative positional encoding that has better generalization and one-to-one correspondence with its absolute counterpart and can be decomposed as:

$$A_{i,j}^{abs} = \underbrace{E_{xi}^T W_q^T W_k E_{xj}}_{(a)} + \underbrace{E_{xi}^T W_q^T W_k U_j}_{(b)}$$
$$+ \underbrace{U_i^T W_q^T W_k E_{xj}}_{(c)} + \underbrace{U_i^T W_q^T W_k U_j}_{(d)} \qquad (6)$$

We provide a novel method for parametrizing the four terms to make this more apparent. This method only uses relative positional data, shown below:

$$A_{i,j}^{rel} = \underbrace{E_{xt}^T W_q^T W_{k,E} E_{x_j}}_{(a)} + \underbrace{E_{x_i}^T W_q^T W_{k,R} R_{i-j}}_{(b)}$$

$$+ \underbrace{u^T W_{k,E} E_{x_j}}_{(c)} + \underbrace{u^T W_{k,R} R_{i-j}}_{(d)} \qquad (7)$$

## 4. Spatial Temporal Relative Transformer Network (ST-RTR)

This architecture leverages Relative Transformer to operate in spatial and temporal networks, which we call ST-RTR. The correlations can be extracted by combining two modules, one for each dimension: Spatial Relative Transformer (S-RTR) and Temporal Relative Transformer (T-RTR). In place of graph convolution, a Transformer is added to skeleton-based action recognition models to address the issues caused by graph convolution's narrow receptive field. We provide a new architecture for transformers called the relative transformer, which preserves the graph's topology while requiring less complexity due to the graph's unordered sequence. The suggested ST-RTR network includes S-RTR stream and T-RTR stream. Another part is ST-GCN feature extraction, which extracts the feature from the previous part and sends it to the following relevant parts: S-RTR and T-RTR. This section will present our ST-RTR model, which uses relative transformers in both spatial and temporal dimensions. The model architecture is depicted in Fig. 2. The S-RTR and the T-RTR are its two component modules. The relative transformer module consists of three node update blocks. These blocks are divided into two sub-blocks: the relay node and the joint node update block. The feedforward neural network (FFN) is associated with both these transformers. Furthermore, the Fusion model takes skeleton joints from the S-RTR stream and T-RTR stream as input, and the spatial-temporal relative transformer computes features as outputs. The output of the two previous steps is combined and then processed by the MLP classification head, and the result is shown in the final action class.

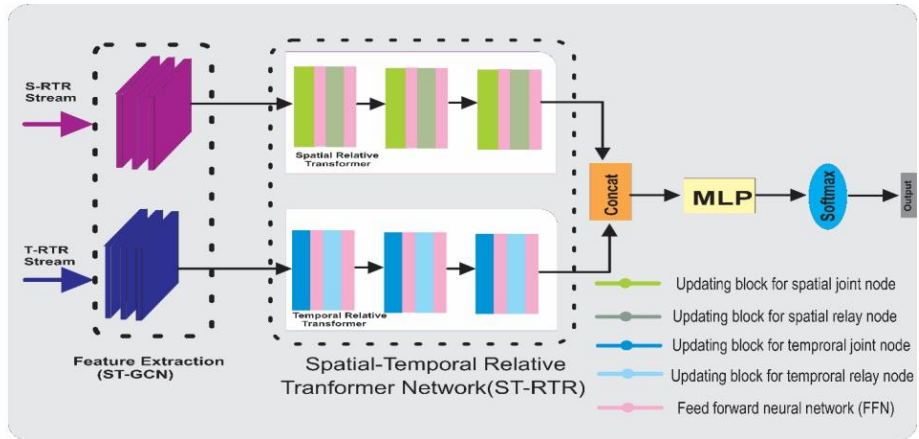

**Fig. 2.** Illustration of two 2s-ST-RTR architectures that comprise S-RTR and T-RTR modules. Three node update blocks are represented by each relative transformer, each of which has two subblocks: joint and relay. Each transformer connects the feedforward neural network (FFN) behind them. The Fusion model inputs skeleton joints from the S-RT and T-RTR streams, and the spatial-temporal relative transformer outputs features. When we concatenate these two outputs, the MLP classification head provides the action class, and SoftMax predicts the final score.

### 4.1 Spatial Relative Transformer (S-RTR)

*Architecture:* We add a virtual node to the skeleton graph as the input, which is different from how the standard transformer works. The spatial-relay node is the virtual node that collects global information from each joint node and sends it to all joint nodes. Two kinds of connections exist between the joint node and the spatial-relay node. These are the inherent connections and the virtual connections.

*Spatial Virtual Connections:* Each joint node and spatial relay node have a connection named after a virtual connection (see Fig. 3(a)). The spatial relay node gets the global composite connection via a virtual connection, giving all joint nodes access to data from non-neighboring joint nodes. A skeleton graph consists of n joint nodes connected by n virtual connections. This structure enables the relative transformer to access local and global information by combining inherent and virtual connections. The standard transformer has inherent and virtual connections. Consequently, with little memory overhead and low computational performance, the model creates a long-range dependency.

*Spatial Inherent Connections:* In Fig. 3(b), we can see that we have created inherent linkages between close joints containing bone connections to maintain the inherent graph topology of skeletons. Because of these connections, each joint node can get local information from the neighboring nodes. However, they allow neighbors to provide more direct communication than non-adjacent joints, supporting the idea that neighbor joints are necessary. A skeleton network with n joint nodes has n-1 intrinsic connections.

*Implementation of S-RTR:* In S-RTR model, we analyze the model's behavior within a single frame. Each frame in the model is equipped with its relative transformer, which we use to examine its performance. The input to the model $Y_{graph} = \{Y_t^1, Y_t^2, ...., Y_t^N\}$ At time $t$, a list of $N$ connected nodes in this frame. $h = (P_{out} + 1/2*\rho*v_{out}^2 + \rho*g*h_{out}) - (P_{in} + 1/2*\rho*v_{in}^2 + \rho*g*h_{in})$ is a collection containing labels for adjacent joint nodes $(R^t)^L$. Each node $Y_x^t$ $(R^t)$ has a query vector $q_x^t$ $(q_r^t)$, a key vector $k_x^t$ $(k_r^t)$, and a value vector $v_x^t$ $(v_r^t)$.

*Block for Spatial Joint Nodes Updating (SJU):* We determine the degree of association between each joint node and its neighbors (containing the neighbor nodes $Y_{B_{y_x^t}}^t$, the relay node $R^t$ and itself $k_v^v$ by doing the dot product of the key and the query vector, as shown in the equation:

$$a_{xy}^t = q_x^t . k_y^{t^T}, x \in N, y \in [x; B_{y_x^t}; r]) \tag{8}$$

Where $a_{xy}^t$ signifies the value of node $y$ on node $x$. To update the joint node, each neighbor node's $v_y^t$ multiplied by its score $a_{xy}^t$ and then sum-up together, as shown below:

$$A_{i,j}^{rel} = \underbrace{E_{xi}^T W_q^T W_{k,E} E_{x_j}}_{(a)} + \underbrace{E_{x_i}^T W_q^T W_{k,R} R_{i-j}}_{(b)}$$
$$+ \underbrace{u^T W_{k,E} E_{x_j}}_{(c)} + \underbrace{u^T W_{k,R} R_{i-j}}_{(d)}$$
(9)

Where $Y_x^t$ is the most recent outcome, combining data from the local and global levels. The key value's channel dimension is $d_k$. (shown in Fig.3(a)). The computations in the model are implemented as the matrix. First, the $q_x^t$ $k_y^t$ $v_y^t$ vectors are filled into $Q^t$, $K^t$, and $V^t$. The matrix $Q^t \in R^{C*1*N}$ contains all joint query vectors for one skeleton. The matrix $k^t \in R^{C*A*N}$ contains all the key vectors, and the matrix $V^t \in R^{C*A*N}$ contains all the value vectors for the adjacent node matrix (introduced in Section 4.3). $C$ represents the feature dimension, $N$ represents the number of joints in a single skeleton, and $A$ represents the maximum number of adjacent nodes. Attention is defined in matrix form as follows:

$$Att(q_v^r, K_v, V_v) = \sum_{x \in A} (SoftMax(\frac{mask(Q^t \otimes k^t)}{\sqrt{d_k}}) \otimes V^t)$$
(10)

Where $\otimes$ is a Hadamard product, the mask removes the padding's zeros.

***Block for Spatial Relay Node Updating (SRU):*** We also use a transformer to ensure the spatial relay node combines the data from all joint nodes better (see Fig. 3(c)). The query vector $q_r^t$ and the key vector $k_y^t$ calculate the significance of each joint node $a_{ry}^t$, as displayed in the following:

$$a_{ry}^t = q_r^t . k_y^{t^T}, y \in [r; N]$$
(11)

The relay node $R^t$ is updated by:

$$R^t = \sum_y SoftMax_y(\frac{a_{ry}^t}{\sqrt{dk}})v_y^t, y \in [r; N]$$
(12)

All the vector values $v_y^t$ in the key vectors $k_y^t$ are put into the matrix key. $V^t \in R^{C*L}$ and $k^t \in R^{C*L}$, respectively. Attention is defined in matrix form as follows:

$$Att(q_r^t, k^t, V^t) = SoftMax(\frac{q_r^t . k^t}{\sqrt{dk}}).(V^t)^T$$
(13)

Where $q_r^t \in R^{1*C}$ denotes the spatial relay node represents the matrix product.

The spatial relative transformer updates joint and relay nodes to gather local and non-local information for an input graph. The algorithm for updating the (S-RTR) is given in Algorithm 1.

---

**Algorithm 1:** *The update of the spatial relative transformer*

**Input:** *Embedded skeleton feature* $Z_1^t$, $Z_2^t$, $\ldots$, $Z_N^t$

**Output :** *The joint nodes* $(Y_N^t)^l$, $(Y_2^t)^L$, $\ldots$, $(Y_N^t)^L$ *and relay node* $(R^t)^L$
*after* L *updates*

**1** // Initialization

**2** $(R^t)^{l-1}$, $(Y_1^t)^0$, $\ldots$, $(Y_N^t)^0$  $Z_1^t$, $Z_2^t$, $\ldots$, $Z_N^t$

**3** $(R^t)^0 \leftarrow average(\mathcal{Z}_1^t\ Z_2^t,...,Z_N^t)$

**4 for** $l \leftarrow 1$ **to** L **do**

**5**      // Update the spatial joint nodes

**6**      **for** x $\leftarrow 1$ **to** N **do**

**7**        $(Y_x^t)^1 = \mathbf{SJU}((Y_1^t)^{l-1}$, $(Y_2^t)^{l-1}$, $\ldots$, $(Y_N^t)^{l-1}$, $(R^t)^{l-1})$

**8**      **end**

**9**      $(Y_1^t)^l$, $(Y_2^t)^l$, $\ldots$, $(Y_N^t)^l = \mathbf{FFN}(Y_1^t)^l$, $(Y_2^t)^l$, $\ldots$, $(Y_N^t)^l)$

**10**     // Update the spatial relay node

**11**     $(R^t)^l = \mathbf{SRU}((R^t)^{l-1}$, $(Y_1^t)^l$, $(Y_2^t)^l$, $\ldots$, $(Y_N^t)^l)$

**12**     $(R^t)^l = \mathbf{FFN}((R^t)^l)$

**13 end**

---

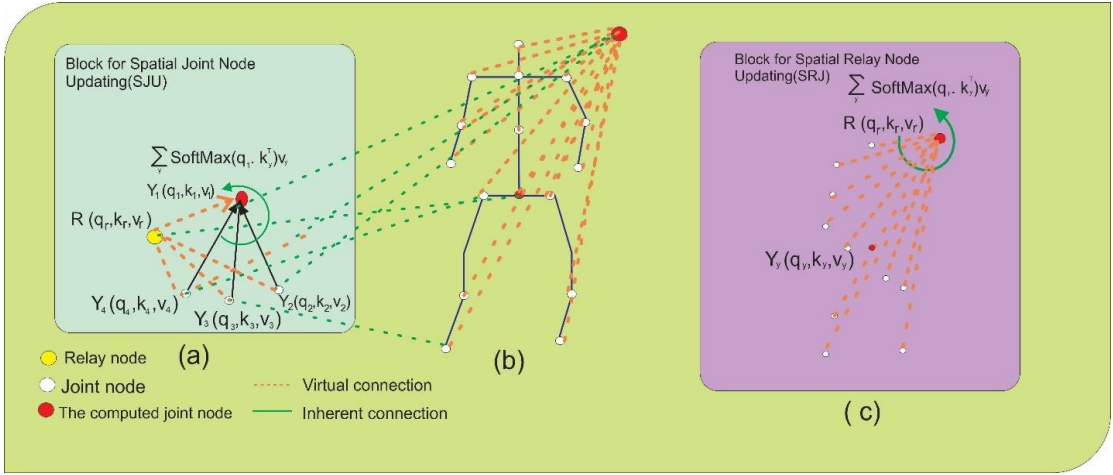

**Fig. 3. (a)** The update process involves getting information from nearby joint nodes and a particular node called the spatial-relay node for each joint node. **(b)** The image illustrates the update process in an S-RTR. The structure of the S-RTR is also presented. **(c)** The spatial relay nodes are updated by evaluating the importance of each node, including the spatial joint and relay node itself.

### 4.2 Temporal Relative Transformer (T-RTR)

After developing an ST-RTR for each skeleton frame, a T-RTR is created for the skeleton sequence. Using a temporal-relay node, the T-RTR has inherent and virtual connections like the S-RTR.

***Temporal Virtual Connections:*** Each virtual connection in the T-RTR, like the operation in the S-RTR, connects a joint node to the temporal relay node. Fig.4b shows a sequence with n nodes with n virtual connections. In short, the temporal relative transformer captures neighboring frame relationships with inherent connections and long-range virtual connections. This separation of semantic compositions into inherent and virtual links allows the model to function without extensive pre-training and decreases the number of connections from n² to 2n, where n represents the skeleton sequence length.

***Temporal Inherent Connections:*** Similar joints in consecutive frames are input to the model as a sequence along the temporal axis. The joints in the first and last frames are also connected to the same joints in the adjacent frames, forming a ring-shaped structure as depicted in Fig.4d. This results in a chain of *n* nodes with *n* inherent connections.

***Implementation of (T-RTR):*** In the T-RTR model, each node is treated as an independent entity. The model is applied to a sequence $Y = \{Y_v^1, Y_v^2, Y_v^t\}$, where each frame represents the same joint node. Each node $Y_v^x$ ($R_v$) consists of a query vector $q_v^x(q_v^r)$, a key vector $k_v^x(k_v^r)$, and a value vector $v_v^x(v_v^r)$.

***Block of Temporal Joint Nodes Updating (TJU):*** As depicted in Fig. 4a, relay node $R_v$ updates the joint node $Y_v^x$, the similar joint node in the adjacent frames $(Y_{x-1}^v, Y_{x+1}^v)$, and itself. The following is an expression of $a_v^{xy}$ score.

$$a_v^{xy} = q_v^x . k_v^{y^T}, x \in T, y \in [x-1; x; x+1; r] \qquad (14)$$

where $a_v^{xy}$ is the importance of the $y^{th}$ frame node on the $x^{th}$ frame node. $Y_v^x$ is the joint note and updated as:

$$Y_v^x = \sum_y SoftMax_j(\frac{a_v^{xy}}{\sqrt{d_k}})v_v^y, x \in T, y \in [x-1; x+1; r] \qquad (15)$$

To represent the matrix key, the query vectors qxv are grouped into a matrix called $Q_v \in R^{c*1*T}$, the key vectors $k_y^v$ are grouped into a matrix called $k_v \in R^{c*B*T}$, and the value vectors $v_y^v$ are grouped into a matrix called $V_v \in R^{c*B*T}$. All key vectors $y$ is represented by variable *B*. The attention means by following matrices:

$$Att(q_v^r, K_v, V_v) = \sum_{x \in A}(SoftMax(\frac{mask(Q^t \otimes K^t)}{\sqrt{dk}}) \otimes V^t) \qquad (16)$$

Where $\otimes$ denotes Hadamard product.

**Block of Temporal Relay Node Updating (TRU):** In Fig. 4c, all the frames' worth of data is sent to the temporal-relay node via scaled dot-product attention, represented as given below:

$$a_v^{ry} = q_v^r . k_v^{y^T}, y \in [r;T] \tag{17}$$

$$R_v = \sum_y SoftMax_y(\frac{a_v^{ry}}{\sqrt{dk}})v_v^y, y \in [r;T] \tag{18}$$

Where $a_v^{ry}$ and $\frac{1}{\sqrt{dk}}$ are attention and scaling score factors, respectively, the relay node is denoted by $R_v$. All key vectors $k_v^v$ derived from $k_v \in R^{e*T}$ and value vectors $v_v^y$ from $V_v \in R^{e*T}$. The attention derived as shown below:

$$Att(q_v^r, K_v, V_v) = SoftMax(\frac{q_v^r . k_v}{\sqrt{dk}}).(V_v)^T \tag{19}$$

Where $q_v^r \in R^{1*c}$ is the temporal relay node, the matrix product is represented by '·'.

The (T-RTR) can capture relations in a sequence of input frames by updating each frame's relay node and the joint node. The algorithm for updating the (T-RTR) is represented in Algorithm 2

---

**Algorithm 2:** *The update of the temporal relative transformer*

**Input:** *Embedded skeleton feature* $Z_v^1, Z_v^2, \cdots, Z_v^T$

**Output:** *The joint nodes* $(Y_v^1)^L, (Y_v^2)^L, \cdots, (Y_v^T)^L$ *and relay node* $(R_v)^L$ *after* L *updates*

**1** // Initialization

**2** $(Y_v^1)^0, (Y_v^2)^0, \cdots, (Y_v^T)^0 \leftarrow Z_v^1, Z_v^2, \cdots, Z_v^T$

**3** $(R_v)^0 \leftarrow$ **average** $(Z_v^1, Z_v^2, \cdots, Z_v^T)$

**4 for** $l \leftarrow$ 1 **to** L **do**

**5**      // Update the temporal joint nodes

**6**      **for** x $\leftarrow l$ **to** T **do**

**7**           $(Y_v^1)^l = $ **TJU**$((Y_v^1)^{l-1}, (Y_v^2)^{l-1}, \cdots, (Y_v^2)^{l-1}, (R_v)^{l-1})$

**8**      **end**

**9**      $(Y_v^1)^l, (Y_v^2)^l, \cdots, (Y_v^T)l = $ **FFN** $(Y_v^1)^l, (Y_v^2)^l, \cdots, (Y_v^T)l$

**10**      // Update the temporal relay node

**11**      $(R_v)^l = $ ***TRU***$((R_v)^{l-1}, (Y_v^1)^l, (Y_v^2)^l, \cdots, (Y_v^T)l$

**12**      $(R_v)l = $ **FFN**$((R_v)^l)$

**13 end**

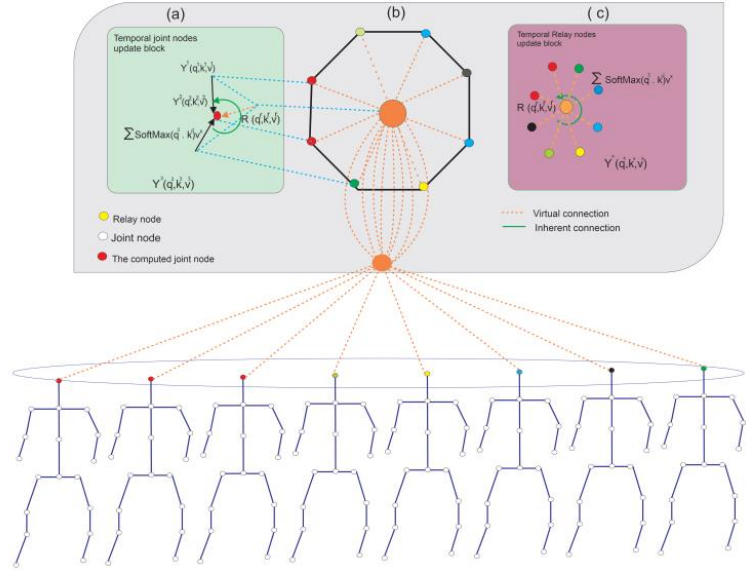

**Fig. 4.** Temporal relative transformer update blocks diagram (T-RTR). (a) TJU modifies both the joint nodes and the temporal relay node. (b) The above approach to doing things is like a ring structure. (c) The impact of each node, with temporal joint and relay nodes, is scored, and the relay node is updated. (d) The joint nodes in the first and final frames and all sampled skeletons are joined sequentially.

### 4.3 Detail of (S-RTR) and (T-RTR)

The multi-head attention mechanism is used in a transformer module S-RTR and T-RTR. The S-RTR module processes an input tensor with dimensions (B, C, V, T); B is the batch size (the number of samples in the input tensor), C is the channel size (the number of channels in the input tensor), V is joint nodes in a skeleton (a structure representing the bones and joints of a body), and T means frames in the sequence (the number of time steps in the input tensor). Fig. 5 describes the S-RTR module. The T and B are switched to obtain a new shape of (B x T, C, V, 1), which blows the transformer to operate independently on each frame. The T-RTR module works similarly to the S-RTR module, but it rearranges the dimensions of the input tensor to have a shape of (B x V, C, T, 1) before processing it. The B and V dimensions are combined: C is the channel size, T represents frames in the sequence, and 1 is a placeholder dimension. Each joint obtains the transformer along the time dimension.

For joints with different nodes, the S-RTR module pads zeros to the nodes with fewer adjacent nodes. This results in an adjoining node matrix of shape (N x A), where joint and adjacent nodes on the skeleton denote N, A. A mask prevents the attention mechanism from considering the padded nodes. The multi-head attention mechanism is then applied to the input matrices Q', K,' and V' using learnable parameter matrices $W_i^Q$, $W_i^K$, and $W_i^V$ for each $head_i$ in the range [1, $N_h$].

The attention mechanism computes a weighted sum of the input values, where the weights are determined by the dot product of the query matrix Q' and the key matrix K' scaled by the inverse of the channel dimension $d_k^i$ for $head_i$. The resulting attention matrix is concatenated with the value matrix V' and multiplied by the parameter matrix $W_i^V$ to obtain the final output for $head_i$. The results from all heads are then attached to get the final multi-head attention output.

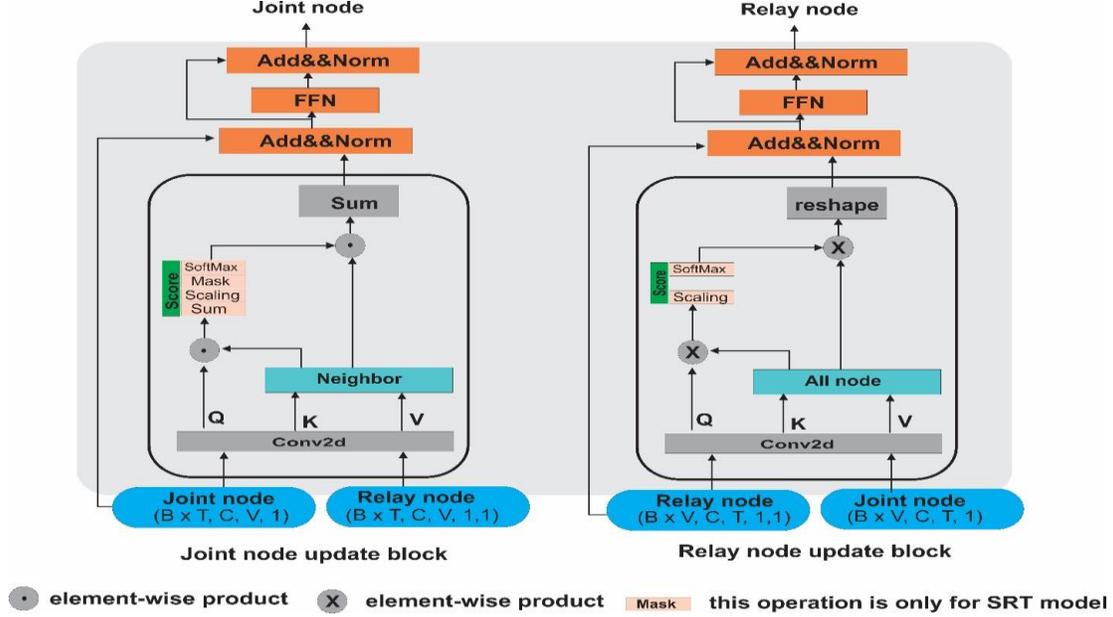

**Fig. 5.** Detailed illustration of the S-RTR module and T-RTR module. The ellipses indicate the S-RTR module and the T-RTR module. The "mask" operation is only employed in S-RTR. According to Fig.2a and 3a, the "neighbor" operation in the ST-RTR module identifies the nodes next to each node.

### 4.4 Transformer-based Fusion Strategy

In the fusion model, the ST-GCN model takes skeleton joints from the S-RTR stream and T-RTR stream as input, and the spatial-temporal relative transformer computes features as outputs. The output of the two previous steps is combined and then processed by the MLP classification head; the result is shown in the final action class. All video frames repeat this process to evaluate cross-frame temporal correlation; we put all frames into spatial-temporal relative transformer blocks. Each token is a joint in the spatial relative transformer, whereas each token is a feature vector expressive one frame for T-RTR.

## 5. Model Evaluation

We extensively test the Spatial and Temporal Transformer streams on NTU-RGB+D 60 to understand their effects [39] (see Table 2). We next compare the optimal configurations to the state-of-the-art on the UAV-Human dataset [40] and the NTURGB+D 120 dataset [41] (see Tables 2–3).

### 5.1 Datasets

Extensive experiments are performed on three RGB-D action recognition datasets to validate the efficacy and significance of our method: NTU RGB+D 60 [39], NTU RGB+D 120 [41], and UAV-Human [40].

***NTU RGB+D 60 (NTU-60):*** (NTU-60) [39] is a 3D human action recognition benchmark dataset, and it was collected by [39] using Microsoft Kinect v2. The skeleton data includes 3D coordinates for 25 body joints and labels for 60 different actions. The NTU-60 dataset is evaluated using two distinct conditions. Based on the camera perspective from which the action was recorded, the Cross-View Evaluation (X-View) dataset has 39,816 training samples, 11,376 validation samples, and 5,688 testing samples. The second, called Cross-Subject Evaluation (X-Sub), comprises 46816 training samples, 13376 validation samples, and 6688 testing samples. Data was collected by having 40 participants do actions and dividing them into three groups, one for training, a second for validation, and a third for testing.

***NTU RGB+D 120 (NTU-120):*** (NTU-120) [41] is an addition of NTU-60, which contains 57,367 new and distinct skeleton sequences, each representing one of 60 different actions. The enlarged dataset uses two criteria to perform the evaluation: The evaluation process consists of two parts: Cross-Subject Evaluation (X-Sub), which is also used for the NTU-60 dataset. Additionally, Cross-Setup Evaluation (X-Set) is used instead of Cross-View. The samples are separated into training, validation, and testing groups based on the parity of the camera setup ID.

***UAV-Human:*** UAV-Human [40] is a novel dataset important for real-world UAV applications. This dataset is made up of distinct human behaviors. A flying UAV composed it day and night in rural and urban areas. The dataset contains 155 actions, classified into six modes: skeleton sequences, RGB, night vision, IR, depth, and fisheye. The dataset for skeleton data recognition includes 22,476 frames with the 2D positions of 17 key points on the human body. These frames are divided into 15,733 for training, 4,495 for validation, and 2,248 for testing.

### 5.2 Implementation Detail

We trained our models on NTU-120 and NTU-60 using the PyTorch [42] framework with a batch size of 32 and SGD as the optimizer for 120 epochs, while for UAV-Human, we used 65 epochs of training and a batch size of 128. Furthermore, we followed the same preprocessing approach as [4, 43]. Drop Attention is a dropout strategy for regularizing attention weights in Transformer networks [44]. To avoid overfitting, it drops the attention logit matrix columns at random. Data augmentation was applied by randomly rotating the 3D skeletons to improve the network's performance and increase the samples' diversity. This method supports the network to learn generalization better and improve its performance. A two-phase training plan was also used to make the model converge faster and more reliably. In the first phase of training (the first 700 steps), the learning rate went from $4 \times 10^{-7}$ to 0.0005 with the gradual warmup strategy. In the second phase, the learning rate went from 0.0005 to 0.9996 with natural exponential decay. We did not conduct a grid search using these criteria. The overall parameter setting for all approaches is given in Table 1.

Regarding the model design, each stream consists of nine layers with channel dimensions of 64,64,64,128,128,128,256,256,256, and 256. The model applies Batch Normalization to the input coordinates, followed by a Global Average Pooling Layer. The final SoftMax classifier is applied, using the standard cross-entropy loss to train each stream.

**Table 1.** Parameter settings for each approach.

| Parameter | NTU-120 | NTU-60 | UAV-Human |
|---|---|---|---|
| Batch Size | 32 | 32 | 128 |
| Optimizer | SGD | SGD | SGD |
| Epochs | 120 | 120 | 65 |
| Training Phases | Two Phases | Two Phases | Two Phases |
| Phase 1: Learning Rate Range | $2e^{-7}$ to 0.0008 | $3e^{-7}$ to 0.0006 | $1e^{-7}$ to 0.0005 |
| Phase 2: Learning Rate Range | 0.0003 to 0.9991 | 0.0004 to0.9985 | 0.0002 to 0.9993 |
| Model Design: Stream Layers | 11 layers | 14 layers | 9 layers |
| Model Design: Batch Normalization | Applied | Applied | Applied |
| Model Design: Global Average Pooling | Applied | Applied | Applied |
| Model Design: SoftMax Classifier | Applied | Applied | Applied |
| Loss Function | Cross-Entropy | Cross-Entropy | Cross-Entropy |

## 5.3. Compared to State-of-the-Art

We evaluate proposed ST-RTR model outcomes with the state-of-art on different datasets like NTU-60, NTU-120, and UAV-Human. Furthermore, we applied the fusion model to the outcomes from the proposed model and got more accurate results, as shown in Tables 2-4.

**Table 2.** NTU-60 comparison Performance.

| Methods | Year | CS % | CV % |
|---|---|---|---|
| HBRNN-L [45] | 2015 | 59.10 | 64.00 |
| Part-Aware LSTM [39] | 2016 | 62.90 | 70.30 |
| ST-LSTM+Trust Gate [46] | 2016 | 69.20 | 77.70 |
| Two-stream RNN [47] | 2017 | 71.30 | 79.50 |
| STA-LSTM [48] | 2017 | 73.40 | 81.20 |
| VA-LSTM [49] | 2017 | 79.40 | 87.60 |
| ST-GCN [3] | 2018 | 81.50 | 88.30 |
| DPRL+GCNN [50] | 2018 | 83.50 | 89.80 |
| HCN [51] | 2018 | 86.50 | 91.90 |
| AS-GCN [20] | 2019 | 86.80 | 94.20 |
| TS-SAN [23] | 2020 | 87.20 | 92.70 |
| MSST-RT [13] | 2021 | 88.43 | 93.21 |
| Fusing+AGE [52] | 2022 | 88.70 | 94.50 |
| ST-RTR | - | 89.10 | 94.90 |
| ST-RTR+Fusion | - | 90.81 | 95.95 |

Table 2 presents that ST-RTR has a good performance, with Cross Subject (CS) and Cross View (CV) settings of NTU RGB+D 60 of 89.10% and 94.90%, respectively, while ST-RTR+Fusion has 90.81% and 95.95% for (CS) and (CV) settings respectively which are higher as compared to the ST-RTR model.

It is noteworthy that both STA-LSTM and MSST-RT utilize the attention mechanism, like the approach proposed in our model. However, STA-LSTM combines the attention mechanism with LSTM, while our model solely relies on the attention mechanism. In terms of performance, our model outperforms STA-LSTM by 15.70% on CS and 13.70% on the (CV). When comparing our model to Fusing+AGE, it should be noted that Fusing+AGE is used to analyze the relationship between video frames and the relationship between the joint nodes of the skeletons. Despite this, our proposed model ST-RTR outperforms Fusing+AGE by 0.40% CS and CV. When we applied fusion to the ST-RTR model, the exceeds were higher than the previous outperforms, and the ST-RTR+Fusion became more potent to the state-of-the-results given in Table 2. After using the fusion, the CS 2.11% and CV 1.45% are higher than our model outperforms.

**Table 3.** NTU-120 performance comparison.

| Method | Year | CS% | CV% |
|---|---|---|---|
| Part-Aware LSTM [39] | 2016 | 25.3 | 26.3 |
| ST-LSTM + Trust Gate [46] | 2016 | 55.7 | 57.9 |
| GCA-LSTM [18] | 2017 | 58.3 | 59.2 |
| Two-Stream GCA-LSTM [17] | 2017 | 61.2 | 63.3 |
| RotClips+MTCNN [53] | 2018 | 64.6 | 66.9 |
| SGN [54] | 2020 | 79.2 | 81.5 |
| MSST-RT [13] | 2021 | 79.33 | 82.30 |
| Fusing+AGE [52] | 2022 | 83.20 | 83.70 |
| ST-RTR | - | 83.50 | 84.10 |
| ST-RTR+Fusion | - | 84.45 | 84.75 |

From Table 3, the suggested ST-RTR model attains the best performance, with 83.50% for the CS and 84.10% for the CV. The proposed model outperforms it by 0.30% for CS and 0.40% for CV. When we applied fusion to our proposed model, the outperforms were higher than the previous outperforms, and our model became more potent to the state-of-the-results presented in Table 3. After using the fusion, the CS 1.25% and CV 1.05% are higher than Fusing+AGE model performance.

**Table 4.** UAV-Human performance comparison.

| Method | Year | Acc. (%) |
|---|---|---|
| ST-GCN [3] | 2018 | 30.25 |
| DGNN [4] | 2019 | 29.90 |
| 2s-AGCN [4] | 2019 | 34.84 |
| HARD-Net [55] | 2020 | 36.97 |
| Shift-GCN [5] | 2020 | 37.98 |
| MSST-RT [13] | 2021 | 41.22 |
| JLF [56] | 2022 | 39.99 |
| ST-RTR | - | 43.01 |
| ST-RTR+Fusion | - | 43.76 |

Table 4 suggests ST-RTR model performs best at 43.01%, outperforming the second-best model by 1.79%. Our model is being compared to the UAV-Human dataset, released in 2021 by [39]. When we applied fusion to our proposed model, the outperforms were higher than the previous outperforms. After using the fusion, the accuracy is 2.54% higher than MSST-RT model.

### 5.4 Performance of Fusion Model

In our fusion model, we use the spatial-temporal relative transformer blocks to compute the temporal correlations between the frames by inputting all the frames into the blocks. Each joint in the dataset is denoted as a token in the S-RTR, and each frame is represented as a feature vector token in the T-RTR. The fusion process improves the model's generalization and performance by combining the features extracted from the S-RTR and T-RTR streams. The fusion model aggregates the spatial and temporal information and captures the correlations between them, resulting in a more accurate and robust model for skeleton-based action recognition. Fig. 6 provides a visual representation of the fusion process and illustrates how it improves the performance of the ST-RTR model at different datasets.

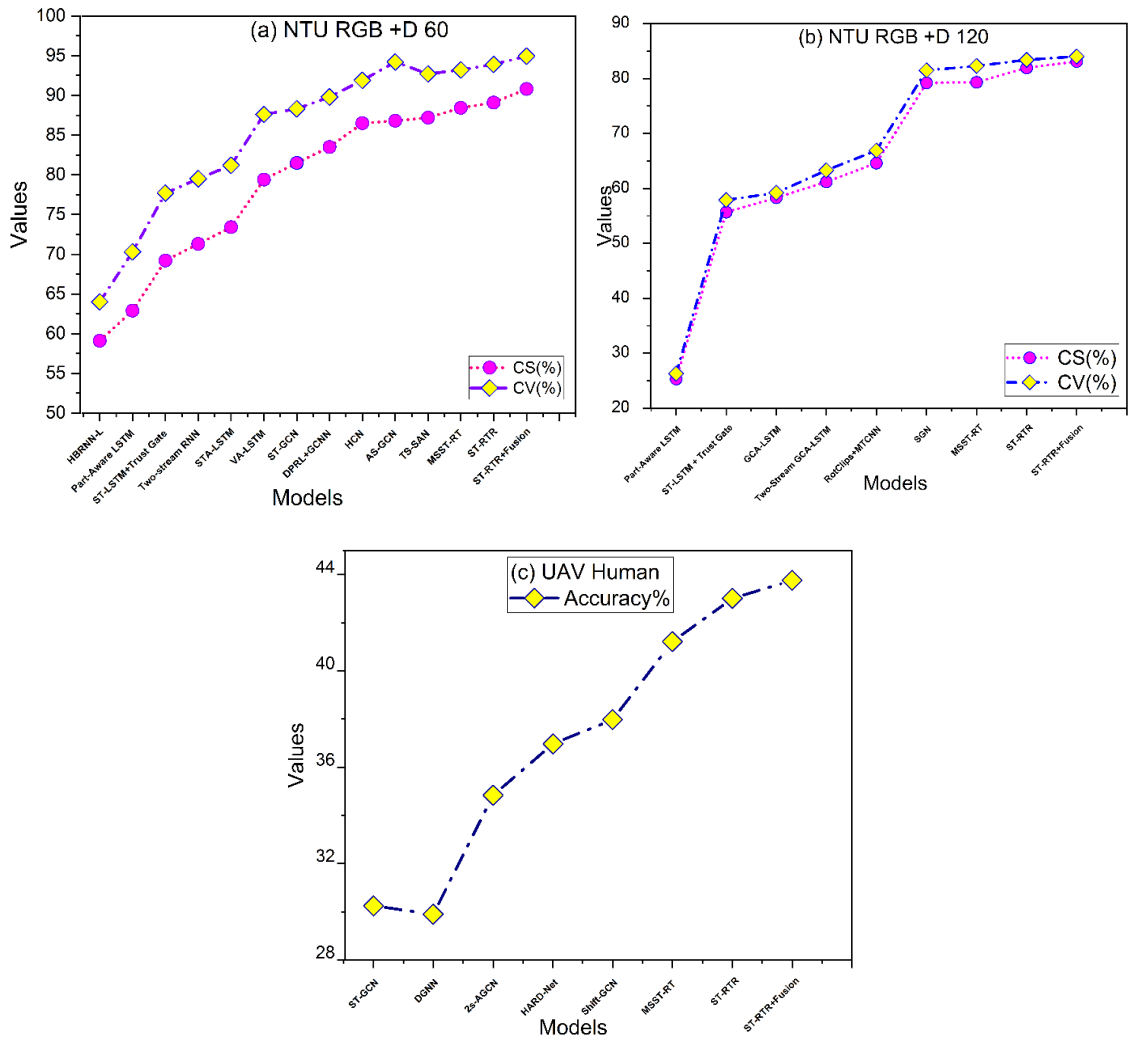

**Fig. 6.** (a) NTU RGB+D 60 Fusion Accuracy. (b) NTU RGB+D 120 Fusion Accuracy. (c) UAV-Human Fusion Accuracy.

### 5.5 Performance Evaluate of Model

In our comprehensive evaluation of the ST-RTR method across the NTU RGB+D 60 and NTU RGB+D 120 benchmarks, we employed a rigorous assessment framework, primarily relying on the confusion matrix as a key metric to measure the model's performance, as shown in Fig. 7. This matrix allowed us to analyze the true positives, true negatives, false positives, and false negatives, providing a nuanced understanding of the classification outcomes. The superiority of our proposed ST-RTR model over the conventional ST-GCN approach was demonstrated through enhanced accuracy, precision, recall, and F1-score values across diverse activities and scenarios. Notably, the ST-RTR model exhibited superior spatiotemporal reasoning capabilities, more effectively capturing intricate dependencies in human actions. Throughout the evaluation process, we encountered challenges related to the inherent complexity of the datasets, such as variations in lighting conditions and diverse human poses. We employed robust preprocessing techniques and augmentation strategies to address these challenges, ensuring the model's robustness and generalizability across varied real-world scenarios. Our findings underscore the efficacy of the ST-RTR method in advancing the state-of-the-art in human action recognition, showcasing its robust performance and potential for real-world applications.

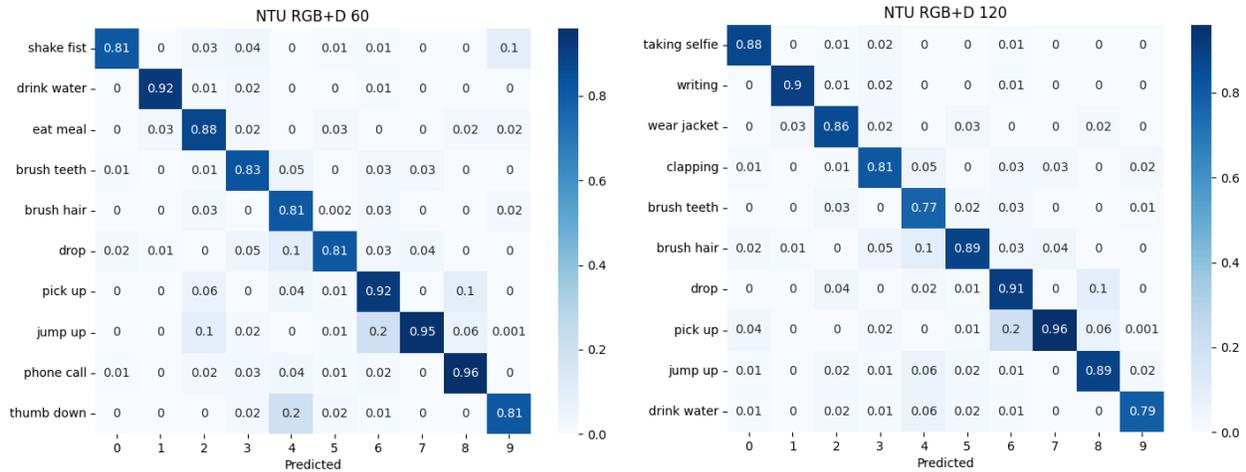

**Fig. 7.** Confusion matrix for performance evaluation of ST-RTR model.

### 5.6 Visualization of SRU and TRU

The proposed model employs an attention mechanism when updating joint and virtual nodes in spatial and temporal dimensions to improve its performance. This approach allows the model to focus on the essential features of the data and improve its ability to classify the action being performed. The attention response can be observed from the last SRU block in the S-RTR and the previous TRU block in the T-RTR. Fig. 8 shows the attention response of 10 multi-heads from the final (SRU) layer during the action of drinking water. The red circles are the nodes for the spatial relay, and the black circles are the nodes for the joint. In the picture, the pink lines show inherent connections, while the green lines show virtual connections. The five nodes with the most attention have been zoomed in, and the other nodes are shown as small circles.

Each attention head has a different response. Head1, Head2, Head3, Head4, and Head5 all pay the most attention to the left hand. This shows that attention functions like human perception. Fig. 9 presents the attention response for the phrase 'Drinking water across ten different multi-head attention mechanisms from the final TRU. The red circles are the temporal relay nodes. The node on the left in the sequence represents the node's state before it was updated, while the node on the right represents its current state after the update was applied. The remaining 18 blue circles are the eighth joint nodes from the 18 sampled frames. The transparency of lines represents the degree of the attention reaction. The darker color indicates a higher response. This demonstrates that information from many frames, including the temporal-relay node itself before it was modified, converged on the node. We can see that head1 and head5 pay a lot of attention to the left red node in the last layer, the temporal relay node.

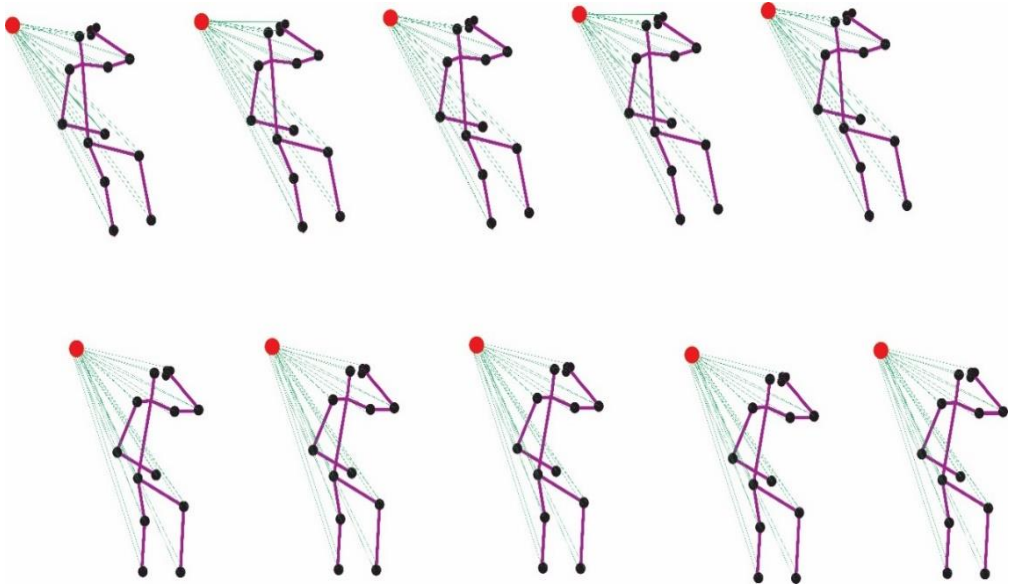

**Fig. 8** Visualization of attention actions during the last (SRU) block in the (ST-RTR) model.

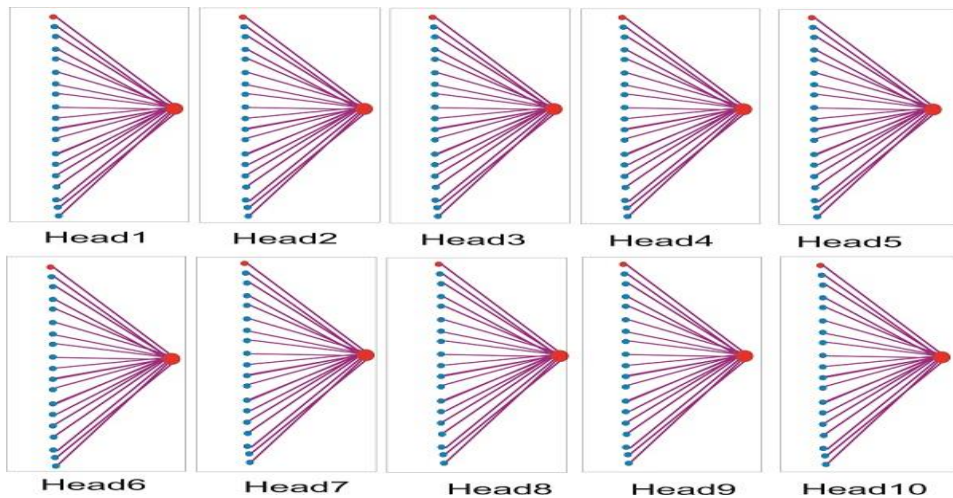

**Fig. 9.** Visualization of the attention action in the last (TRU) block of the (ST-RTR) model.

## 6. Conclusion

This paper proposes a new approach for skeleton action recognition that uses Transformer self-attention as an alternative to graph convolution, particularly in dealing with long-range correlations due to their restricted receptive fields, such as human action in the field of disabled folk and elder homes. This makes it hard to understand human behavior based on long-distance relationships. To tackle this, we introduce a spatial-temporal relative transformer (ST-RTR) model with a relay node for calculating each node's score. ST-RTR breaks the inherent skeleton topology in spatial and the order of skeleton sequence in temporal dimensions. Our proposed architecture, the "relative transformer," is based on the standard transformer but compensates for its limitations while preserving the skeleton's inherent structure and reducing computational complexity. Additionally, this architecture allows for the use of the relative transformer without requiring extensive pre-training. We also modify the relative transformer to create an S-RTR and a T-RTR, which can extract spatial-temporal features. Our final ST-RTR network performs very well on multiple datasets compared to other methods that use similar input joint information and stream setups. When bone information is added to the input used by ST-RTR network, the network performs as effectively as the best methods in the field. However, our comprehensive experiments on NTU-60, NTU120, and UAV-Human explained that ST-RTR model showed that configurations using only self-attention modules did not perform as well.

In future research, a promising avenue is exploring a unified transformer architecture to replace graph convolution across diverse tasks. Addressing a limitation in current human action recognition datasets (NTU-60 and NTU-120), there's a need to focus on detailed hand and finger movements. A potential enhancement involves adopting an alternative method to extract skeleton data from subtle hand movements. This novel approach could significantly augment the performance of ST-RTR by integrating valuable information derived from intricate hand movements, thus advancing the accuracy and comprehensiveness of action recognition systems.

**Declarations**

- **Competing Interests:** The authors declare that they have no known competing financial interests or personal relationships that could have appeared to influence the work reported in this paper.
- **Authors' contributions:** Methodology and formal analysis done by Faisal Mehmood and Touqeer Abbas; Project administration and Data curation done by Enqing Chen; and Samah M. Alzanin does the final revision and English polishing. All authors have read and agreed to the published version of the manuscript.
- **Ethical Approval:** All procedures performed in studies were by the ethical standards of the institutional and national research committee.
- **Availability of data and materials:** Not applicable.
- **Acknowledgments:** The authors thank the King Salman Center for Disability Research for funding this work through Research Group no KSRG-2023-544.